\documentclass[9pt]{article}
\usepackage{amsmath}
\usepackage{graphicx}
\usepackage{amsfonts}
\usepackage{subcaption}
\usepackage{dblfloatfix}
\usepackage{algorithm,algorithmic}
\usepackage[preprint]{spconf}

\newcommand{\argmin}[1]{\underset{#1}{\operatorname{arg}\,\operatorname{min}}\;} 

\title{APPEARANCE-BASED GESTURE RECOGNITION IN THE COMPRESSED DOMAIN}
\name{Shaojie Xu \qquad Anvesha Amaravati \qquad Justin Romberg \qquad Arijit Raychowdhury
\thanks{This research project was supported by a grant from Intel.}
\thanks{Copyright 2017 IEEE. Published in the IEEE 2017 International Conference on Acoustics, Speech, and Signal Processing (ICASSP 2017), scheduled for 5-9 March 2017 in New Orleans, Louisiana, USA. Personal use of this material is permitted. However, permission to reprint/republish this material for advertising or promotional purposes or for creating new collective works for resale or redistribution to servers or lists, or to reuse any copyrighted component of this work in other works, must be obtained from the IEEE. Contact: Manager, Copyrights and Permissions / IEEE Service Center / 445 Hoes Lane / P.O. Box 1331 / Piscataway, NJ 08855-1331, USA. Telephone: + Intl. 908-562-3966.}
}
\address{School of Electrical and Computer Engineering, Georgia Institute of Technology, Atlanta, GA, 30332\\
$\{$kyle.xu, aamaravati3$\}$@gatech.edu, $\{$jrom, arijit.raychowdhury$\}$@ece.gatech.edu}
\begin{document}
%
\maketitle
\begin{abstract}
We propose a novel appearance-based gesture recognition algorithm using compressed domain signal processing techniques. Gesture features are extracted directly from the compressed measurements, which are the block averages and the coded linear combinations of the image sensor's pixel values. We also improve both the computational efficiency and the memory requirement of the previous DTW-based K-NN gesture classifiers. Both simulation testing and hardware implementation strongly support the proposed algorithm.
\end{abstract}

\section{INTRODUCTION AND RELATED WORK}
\label{sec:intro}

Hand gesture recognition is continuously evolving in how systems on chip (SoCs) interact with users. To achieve power efficiency, these SoCs turn into idle mode when not being used and ”wake up” when users are detected. The detection of an user's present requires the sensor front-ends to be perpetually ON, thus making low power consumption an important design criteria. As cameras have become default devices embedded in many systems, a camera-based hand gesture recognition system is suitable for providing stimulus for system wake up. 

Based on image outputs from a camera, most existing algorithms work directly in the pixel domain \cite{pavlovic1997visual, rautaray2015vision}. The majority of the work can be divided into three stages. First, the hand region is extracted from the image using techniques such as background extraction, skin color detection, and contour detection. Second, the motion of the gesture is characterized by features. The common types of features include difference image, motion centroid, optical flow, and motion vectors. At the last stage, these features are sent to a classifier. Dynamic time warping (DTW) with K nearest neighbors (K-NN), hidden Markov models, and neural networks have all been implemented and showed promising result.

Aforementioned algorithms require a significant amount of energy in A/D conversion of each pixel of the image sensor. Reducing the number of sensing measurements plays an important role of energy saving. Recent development in compressive sensing and target recognition in the compressed domain \cite{davenport2007smashed, mantzel2012compressive, duarte2008single} improved the performance and energy efficiency of the overall process of data acquisition, feature extraction and recognition. These works suggest us taking coded combinations of the pixel vales and characterizing the gesture motion directly from a few compressed measurements.

In this paper, we propose an appearance-based gesture recognition algorithm for system wake up. The gesture motion is captured by a sequence of difference images. Each difference image passes through two layers of compression to reduce its resolution and to be transferred to the compressed domain. The parameters of the motion are then directly extracted from the compressed domain and used as features for classification. To the authors' knowledge, our work is the first in gesture recognition using compressive sensing techniques. We also enhance the previous DTW-based K-NN classifiers \cite{ten2007multi, akl2010accelerometer}, which allows them to cooperate with clustering and dimension reduction techniques, and therefore, improves both the computational and the memory efficiency of time series classification. In a hardware co-design, we implemented the compression in the sensor front-end and motion parameter estimation in the mixed signal domain. The testing results show significant energy saving over previous works \cite{shi2007fpga, li2012novel}.

\section{ALGORITHMS}
\label{sec:algo}
\begin{figure}[htb]
\centering
\includegraphics[scale=.4]{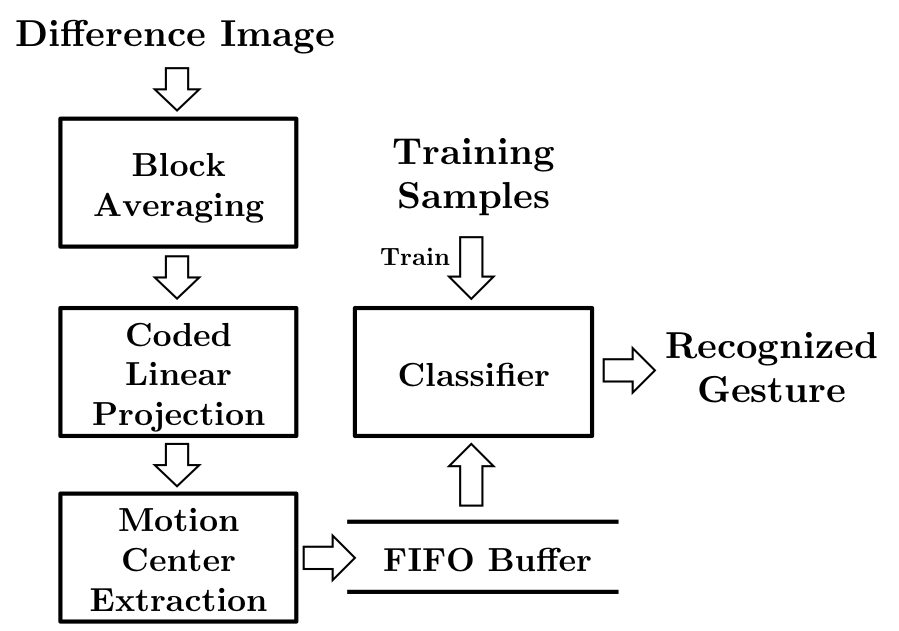} 
\caption{\small Block diagram of the proposed algorithm} \label{Block_Diagram}
\vspace{-.5cm}
\end{figure}

Difference images are capable of capturing gestures containing significant motions. We pass each difference image through two layers of compression. In the first layer, the resolution is reduced by dividing the whole image into several blocks and taking the average of each block. In the second layer, we take coded combinations of these block-averaged pixels. We estimate the center of the motion directly from these compressed measurements. These motion centers are passed to a classifier for gesture recognition. Figure \ref{Block_Diagram} shows the block diagram of our system.

\subsection{Two layers of compression}

\begin{figure}[htb]
\centering
\includegraphics[scale=.55]{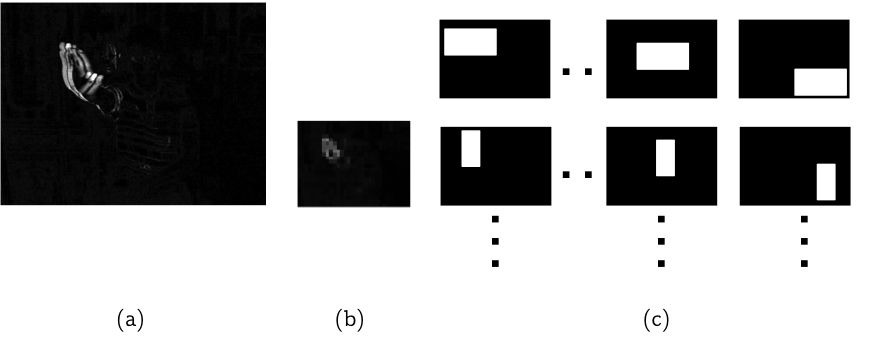}
\caption{\small (a) Full-resolution difference image $D_i$; (b) Block averaged difference image; (c) Matching templates in the uncompressed domain. Rectangle sizes differ among rows and centers of the rectangles differ among columns.} \label{Di_DiBA_X}
\vspace{-.5cm}
\end{figure}

Denote $F_i$ as the $i$th full resolution image output from the camera of size $W\times H$. The difference image $D_i$ (Figure \ref{Di_DiBA_X}.a) of two consecutive frames is calculated as $ D_i=|F_{i+1}-F_i| $.

In the first compression layer, the difference image is divided evenly into blocks of size $B$ by $B$. The average of the pixel values in each block is taken, resulting in a block-compressed difference image of size $W/B$ by $H/B$ (Figure \ref{Di_DiBA_X}.b). We vectorize this low-resolution difference image and denote it as $y_i \in \mathbb{R}^N$. 

In the second layer of compression, we chose a random matrix $\Phi$ of size $M$ by $N$ as the coded measuring matrix. Each entry of $\Phi$ is uniformly chosen from $\{+1, -1\}$. The projection of the vectorized low-resolution difference image in the compressed domain is calculated as:
\begin{equation}
\hat{y}_i = \Phi y_i = \Phi\Psi Y_i \label{eq:cf}
\end{equation}
Each entry in $\hat{y}_i \in \mathbb{R}^M$ is a random linear combination of all the entries in $y_i$. $Y_i$ is the vectorized original difference image $D_i$. $\Psi$ is the block averaging matrix of size $N$ by $W\times H$.

\subsection{Motion center extraction in the compressed domain}
In the uncompressed low-resolution domain, the hand region in the difference image can be captured by a template shown in Figure \ref{Di_DiBA_X}.c. The template (of size $W/B$ by $H/B$) has uniform non-zero values within the small rectangular region and zeros elsewhere. To locate the hand region, we construct a set of vectorized templates $X(\alpha , r)$, where $\alpha$ represents the coordinates of the center of the small rectangle, and $r$ represents different rectangle sizes. The variation in sizes is to adapt to the change of the hand size seen by the camera when users at different locations. The center of the hand motion is extracted by solving
\begin{equation}
(\alpha^*, r^*) = \argmin{\alpha, r} ||y_i - X(\alpha , r)||_2 \label{eq:uc}
\vspace{-.25cm}
\end{equation}
The collection of templates forms a manifold in $\mathbb{R}^N$ with intrinsic parameters $\alpha$ and $r$. Using the result from \cite{davenport2007smashed} and \cite{baraniuk2009random}, we can directly extract the motion centers in the compressed domain. That is, for
\begin{equation}
(\hat{\alpha}^*, \hat{r}^*) = \argmin{\alpha, r} ||\hat{y}_i - \Phi X(\alpha , r)||_2 \label{eq:c1}
\vspace{-.25cm}
\end{equation}
$(\hat{\alpha}^*, \hat{r}^*) \approx (\alpha^*, r^*)$ with high probability for some $M \ll N$. The block averaging layer reduces the possible choices of $r$, and techniques such as matched filtering can be applied to efficiently solve equation (\ref{eq:c1}).

\subsection{Train the gesture classifier}
DTW-based classifiers perform well for dataset containing limited amount of samples \cite{carmona2012performance}. Traditional DTW-based classifiers use DTW \cite{muller2007dynamic} as the distance measuring method between two sequences of different lengths, and use K-NN method for classification. The memory and computational requirements thus grow linearly with the size of training set.

To reduce the number of DTW calculations in the recognition stage, we perform K-means clustering in the training dataset to form ``super samples''. The distance between an individual sample and a super sample is measured using DTW. In each iteration, the super samples are updated as the average of all the samples within their clusters. DTW barycenter averaging (DBA) \cite{petitjean2011global} is used as the averaging method.

The main difficulty of time series classification comes from the different lengths of the samples. We notice that DBA can find clustering centers of an arbitrary length set by the user. The pairwise matching information in DTW also provides a way to rescale the length of time sequences. Therefore, we propose a DTW length rescale algorithm, shown in Algorithm \ref{algo:DLR}:
\begin{algorithm}
\caption{DTW Length Rescaling} \label{algo:DLR}
\begin{algorithmic}
	\small
  	\REQUIRE $K$ ``super samples'' $\mathbf{S_1}, \mathbf{S_2}, \ldots, \mathbf{S_K}$ of length $\tau$, calculated using K-means with DTW and DBA.
  	\REQUIRE Sequence $\mathbf{T}$ to be rescaled to length $\tau$
  	\STATE $\mathbf{S^*} = \argmin{} DTW \left( \mathbf{S_k},\mathbf{T} \right)$
  	\STATE $\mathbb{M} \leftarrow$ pairwise matching information between $\mathbf{S^*}$ and $\mathbf{T}$
  	\STATE Initialize $\mathbf{T}'$ of length $\tau$
  	\FOR{$i=$ 1 to $\tau$}
  	\IF{$S^*_i$ is matched to multiple points $T_{j}, T_{j+1}, \ldots, T_{j+m}$, according to $\mathbb{M}$, }
  	\STATE $T_i' = \argmin{T_l \in  T_{j}, \ldots, T_{j+m}} ||S_i - T_l||$ 
  	\ELSE
  	\STATE $T_i' = T_j$, where $T_j$ is the only matching point to $S^*_i$
  	\ENDIF
  	\ENDFOR
\end{algorithmic}
\end{algorithm}

\begin{figure}[htb]
\centering
\includegraphics[scale=.375]{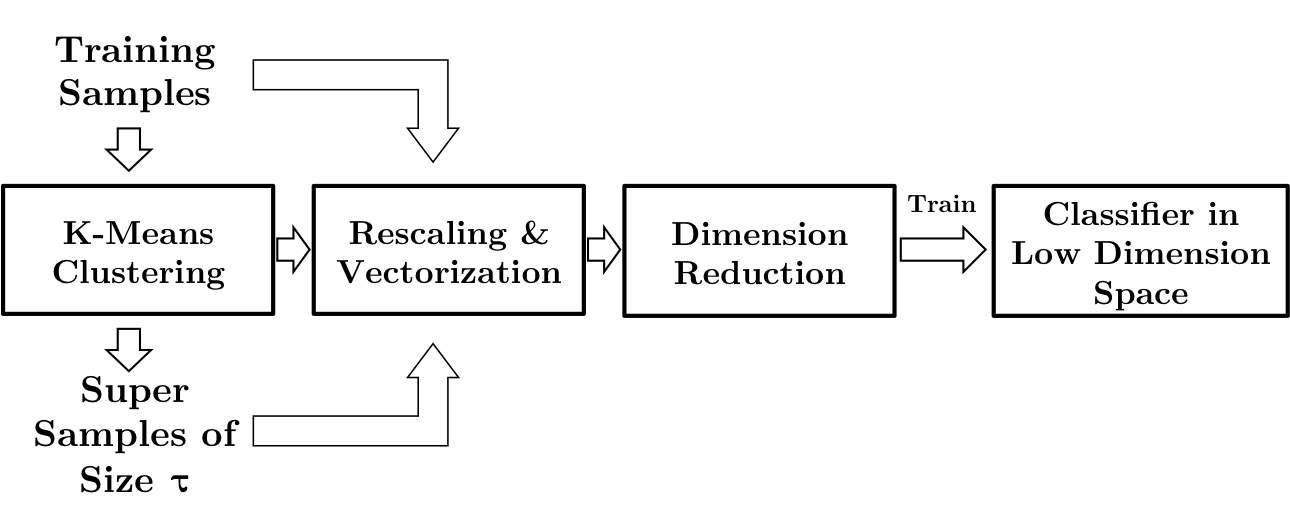}
\caption{\small Block diagram of training the gesture calssifier} \label{Fig:Train_Block}
\vspace{-.5cm}
\end{figure}

Using Algorithm \ref{algo:DLR}, we rescale all the training sequences to the same length $\tau$. Since each motion center in the time sequence contains both x and y coordinates, each gesture sample is now of size $2 \times \tau$. We vectorize the samples by cascading all the y coordinates after the x coordinates, transferring the time series classification problem to a traditional classification problem in $\mathbb{R}^{2\tau}$. Various dimension reduction techniques and multi-class classification algorithms can then be implemented. The block diagram of the complete training procedure is shown in Figure \ref{Fig:Train_Block}.

\subsection{Gesture recognition}
In a real-time system, the extracted motion centers are stored in a FIFO buffer of length $L$. To recognize the gesture, we first rescale the buffer data sequence based on the learned super samples using Algorithm \ref{algo:DLR}. Within this step, open-ended DTW is used for pairwise matching in order to automatically separate the gesture-like data from the noise at both ends of the buffer. The new gesture-like sequence of length $\tau$ then goes through vectorization and dimension reduction before being sent to the trained classifier. 

Compared to the traditional DTW-based K-NN classifiers, our classifier significantly reduces the number of DTW calculation in the recognition stage and still being able to exploit the structure of the entire training set. In our experiments, most of the gestures can be well separated in a very low dimension, making the dimension reduction and the low-dimensional classifier computationally efficient as well.

\section{TESTING AND RESULTS}
\label{sec:testing}
\subsection{Number of compressed measurements}

\begin{figure}[htb]
\centering
\includegraphics[scale=.5]{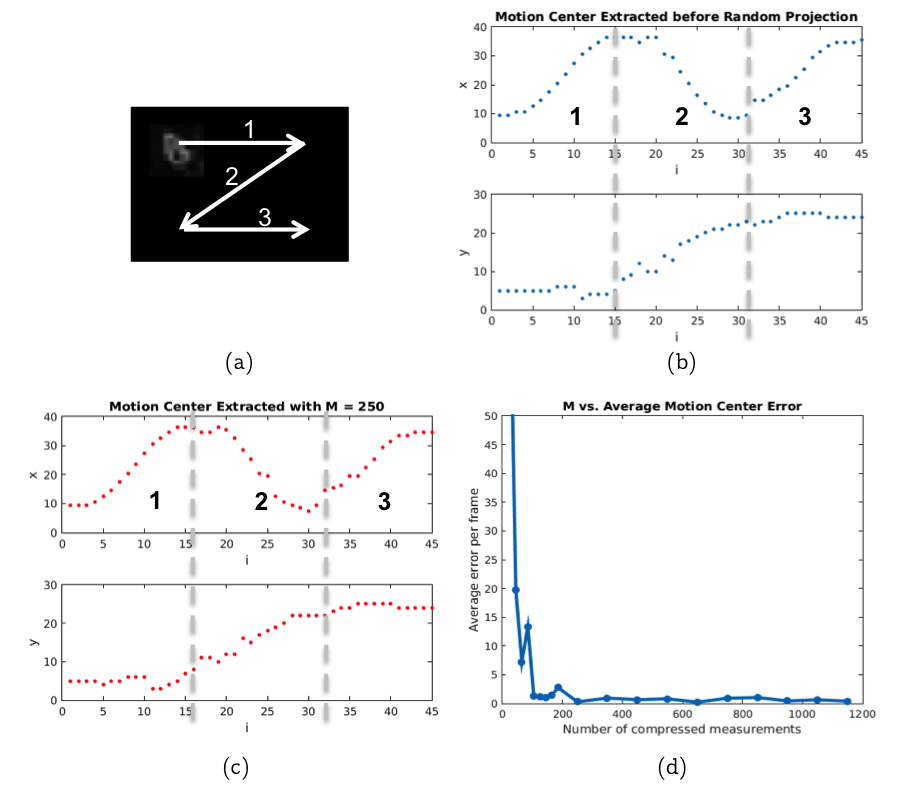}
\caption{\small MATLAB simulation results for different numbers of compressed measurements $M$. (a) Hand motion of gesture "Z" diveded into three segments; (b) Motion center extracted before random projection by solving equation (\ref{eq:uc}); (c) Motion center extracted from 250 compressed measurements by solving equation (\ref{eq:c1}); (d) Average error of motion center extracted in the compressed domain compared with (b).} \label{M_vs_MC}
\vspace{-.5cm}
\end{figure}

To gain better insight for the choice of the number of compressed measurement $M$, we first explored its relationship with the accuracy of motion center extraction. We ran the extraction algorithm with one video of gesture "Z" (Figure \ref{M_vs_MC}.a). In the block averaging layer, difference images of size $480\times 640$ are compressed by blocks of size $16 \times 16$. We extracted the motion centers from these block-averaged difference images by solving equation (\ref{eq:uc}). Shown in Figure \ref{M_vs_MC}.b, the three segments of the gesture are clearly distinguished on the path of the motion centers. This result was used as the ground truth for comparing $M$.

We then used only 250 compressed measurements ($M = 250$) and the motion centers were extracted by solving equation (\ref{eq:c1}), and were plotted in Figure \ref{M_vs_MC}.c. The apparent similarity between this plot and \ref{M_vs_MC}.b verifies the theory. For each value of $M$, we calculated the average motion center error per frame in the compressed domain. The "L" shape of the curve indicates that $M=250$ is the threshold for nearly error-free motion parameter estimation, granting us another factor of 5 compression rate. This "threshold" behavior is consistent with the classic results from compressed sensing presented in \cite{davenport2007smashed,mantzel2012compressive,baraniuk2009random}. The accurate motion center extraction in the compressed domain provides the foundation of high recognition rate.

\subsection{SKIG dataset}
We tested the overall algorithm on the public-available SKIG dataset \cite{liu2013learning}. We selected 5 classes containing significant motion in the x-y plane: Circle, Triangle, Wave, Z, and Cross. In each class, we selected 70 well-illuminated samples and randomly divided them into training (55 samples) and testing (15 samples) sets. We cropped the training videos so that the gesture motion filled the entire video. Since each frame is of size $240 \times 320$, in the block compression layer, we used blocks of size $10$ by $10$, and we set $M=200$ in the random projection layer.

\begin{figure}[htb]
\centering
\includegraphics[scale=.425]{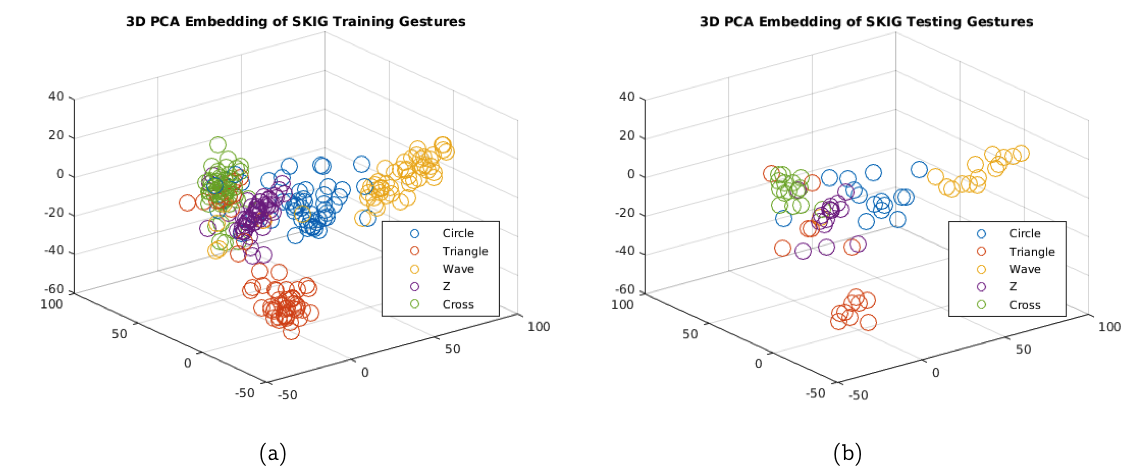}
\caption{\small (a) PCA embedding of SKIG training samples in $\mathbb{R}^3$; (b) PCA embedding of SKIG testing samples in $\mathbb{R}^3$.} \label{Fig:SKIG_PCA}
\end{figure}

During the training stage, we calculated 1 super sample in each gesture class. As the lengths of the gesture videos vary from $48$ to $236$ frame, we chose $\tau$ to be the average length $116$. After rescaling and vectorizing all the training sample, we applied PCA and used the first 3 principle components. The gesture samples were well separated in $\mathbb{R}^3$, as shown in Figure \ref{Fig:SKIG_PCA}.a. For simplicity, we modeled each gesture class' distribution as a multivariate Gaussian. A gesture was assigned to the class with highest likelihood beyond a rejecting threshold. The resulting classifier had decision boundaries of ellipsoid shapes.

Following the proposed recognition procedure, we show the testing samples' PCA embedding in Figure \ref{Fig:SKIG_PCA}.b. The recognition rate is shown in Table \ref{Table:SKIG}. The relatively low recognition rate for triangle gestures is caused by the longer sample videos that usually contain more than $200$ frames. Rescaling them resulted in significant downsampling, and some of these downsampled data overlap with the "Z" gesture samples in the $\mathbb{R}^3$ embedding. Our classifier had comparable performance with a 5-NN classifier for all other classes.

\begin{table}[htb]
\footnotesize
\begin{center}
 \begin{tabular}{|c| c| c| c| c| c|}  
 \hline
 Gesture Type & Circle & Triangle & Wave & Z & Cross \\ 
 \hline
 \shortstack{ \\ Recognition Rate \\ (Our Classifier)} & 93.3\% & 73.3\% & 93.3\% & 80\% & 93.3\%\\ 
 \hline
 \shortstack{ \\ Recognition Rate \\ (DTW 5-NN)} & 93.3\% & 93.3\% & 93.3\% & 100\% & 100\%\\ 
 \hline
\end{tabular}
\end{center}
\vspace{-.25cm}
\caption{\small Recognition rates of SKIG gesture dataset}
\label{Table:SKIG}
\vspace{-.5cm}
\end{table}

\subsection{Real-time OpenCV simulation}
\begin{figure}[htb]
\centering
\includegraphics[scale=.425]{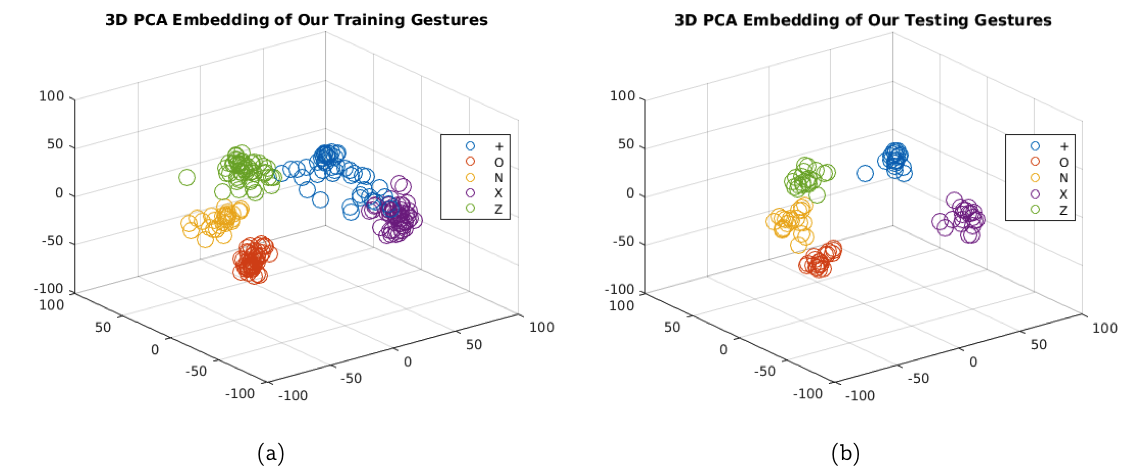}
\caption{\small (a) PCA embedding of our training samples in $\mathbb{R}^3$; (b) PCA embedding of our testing samples in $\mathbb{R}^3$.} \label{Fig:Our_PCA}
\end{figure}

We simulated a real-time system using OpenCV and a webcam as the image sensor. \footnote{Our OpenCV (C++) demo code is available at www.kylexu.net/cs-gesture-recog} Each frame was of size $480 \times 640$. In the block compression layer, we used blocks of size $20$ by $20$ and set $M=200$ in the random projection layer. We specified 5 different gesture classes: ``+'', ``O'', ``N'', ``X'', and ``Z'', each containing 50 samples. Following the similar training procedure as performed on the SKIG dataset, we transferred each gesture sample into a point in $\mathbb{R}^3$, plotted in Figure \ref{Fig:Our_PCA}.a, and trained a Gaussian maximum likelihood classifier.

To calculate the recognition rate, we performed 20 gestures per class in front of the webcam. These testing samples' PCA embedding in $\mathbb{R}^3$ is shown in Figure \ref{Fig:Our_PCA}.b. We calculated the false detection rate by performing 50 unspecified gestures. Table \ref{Table:Our} shows both the recognition rate and the false detection rate. The simulation result strongly verified our algorithm.

\begin{table}[htb]
\footnotesize
\begin{center}
 \begin{tabular}{|c| c| c| c| c| c|}  
 \hline
 Gesture Type & + & O & N & X & Z \\ 
 \hline
 Recognition Rate & 100\% & 100\% & 100\% & 95\% & 100\%\\ 
 \hline
 False Detection Rate & 4\% & 0\% & 4\% & 2\% & 0\%\\ 
 \hline
\end{tabular}
\end{center}
\vspace{-.25cm}
\caption{\small Real-time OpenCV simulation results}
\vspace{-.5cm}
\label{Table:Our}
\end{table}

\subsection{Hardware implementation}
\begin{figure}[htb]
\centering
\includegraphics[scale=.45]{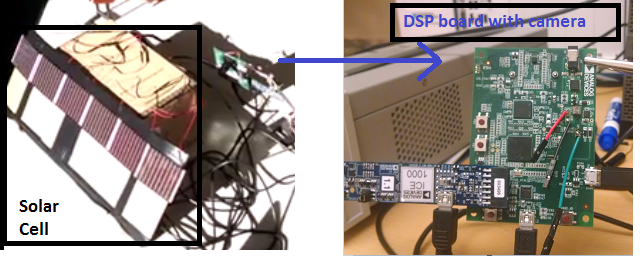}
\caption{\small Our light-powered smart camera system.} \label{Fig:Board_Full}
\vspace{-.5cm}
\end{figure}

We implemnted the proposed algorithm in a gesture recognition system powered by solar energy, shown in Figure \ref{Fig:Board_Full}. With 400 compressed measurement, this system achieved greater than $80\%$ accuracy and consumed only $95mJ$ of energy per frame \cite{anvesha2016light}. In this system, the random projection is simulated in the MCU and the classifier used DTW-based 1-NN method. In an ongoing project, we combined both compression layers into the camera front-end, and implemented motion center extraction in the mixed signal domain. Early testing results have shown the energy consumption reduced to $1.3\mu J$ per frame.

\section{CONCLUSION}
We proposed an energy-efficient appearance-based gesture recognition algorithm in the compressed domain. The major saving of power comes from the two layers of compression that reduce the resolution of the image sensor by a factor of more than $1000$. Our proposed gesture classifier significantly reduces the number of DTW calculations and the memory requirements in the traditional DTW-based K-NN classifiers while preserving the structure of the full training dataset. Our algorithm has been verified by both simulation testing and hardware co-design.

\label{sec:conclusion}


\bibliographystyle{IEEEbib}
\bibliography{CS_Gesture_Recog}

\end{document}